\documentclass[10pt,twocolumn,letterpaper]{article}

\usepackage{cvpr}              
\usepackage{subcaption}
\usepackage{multirow}
\usepackage{colortbl}
\usepackage{pifont}
\usepackage{makecell}
\definecolor{cvprblue}{rgb}{0.21,0.49,0.74}
\usepackage[pagebackref,breaklinks,colorlinks,allcolors=cvprblue]{hyperref}

\def\paperID{6205} 
\def\confName{CVPR}
\def\confYear{2025}

\title{QMamba: Post-Training Quantization for Vision State Space Models}

\author{Yinglong Li \and Xiaoyu Liu
\and
Jiacheng Li \and Ruikang Xu \and Yinda Chen
\and
Zhiwei Xiong
\and
University of Science and Technology of China
}

\begin{document}
\maketitle
\begin{abstract}

State Space Models (SSMs), as key components of Mamaba, have gained increasing attention for vision models recently, thanks to their efficient long sequence modeling capability. 
Given the computational cost of deploying SSMs on resource-limited edge devices, Post-Training Quantization (PTQ) is a technique with the potential for efficient deployment of SSMs.
In this work, we propose QMamba, one of the first PTQ frameworks to our knowledge, designed for vision SSMs based on the analysis of the activation distributions in SSMs. We reveal that the distribution of discrete parameters exhibits long-tailed skewness and the distribution of the hidden state sequence exhibits highly dynamic variations. Correspondingly, we design Long-tailed Skewness Quantization (LtSQ) to quantize discrete parameters and Temporal Group Quantization (TGQ) to quantize hidden states, which reduces the quantization errors.
Extensive experiments demonstrate that QMamba outperforms advanced PTQ methods on vision models across multiple model sizes and architectures. Notably, QMamba surpasses existing methods by 21.0\% on ImageNet classification with 4-bit activations.

\end{abstract}    
\section{Introduction}
\label{sec:intro}

\begin{figure}[t]
  \centering
  \subcaptionbox{Discrete parameters $\overline{A}_t$\label{teaser:deltaA}}{
    \includegraphics[width=0.475\linewidth]{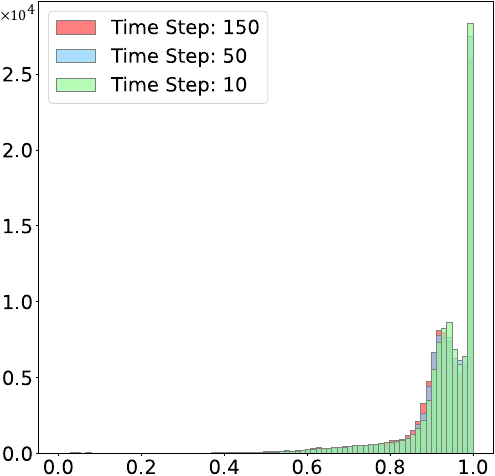}
  }\hfill 
  \subcaptionbox{Hidden states $h_t$ \label{teaser:state}}{
    \includegraphics[width=0.48\linewidth]{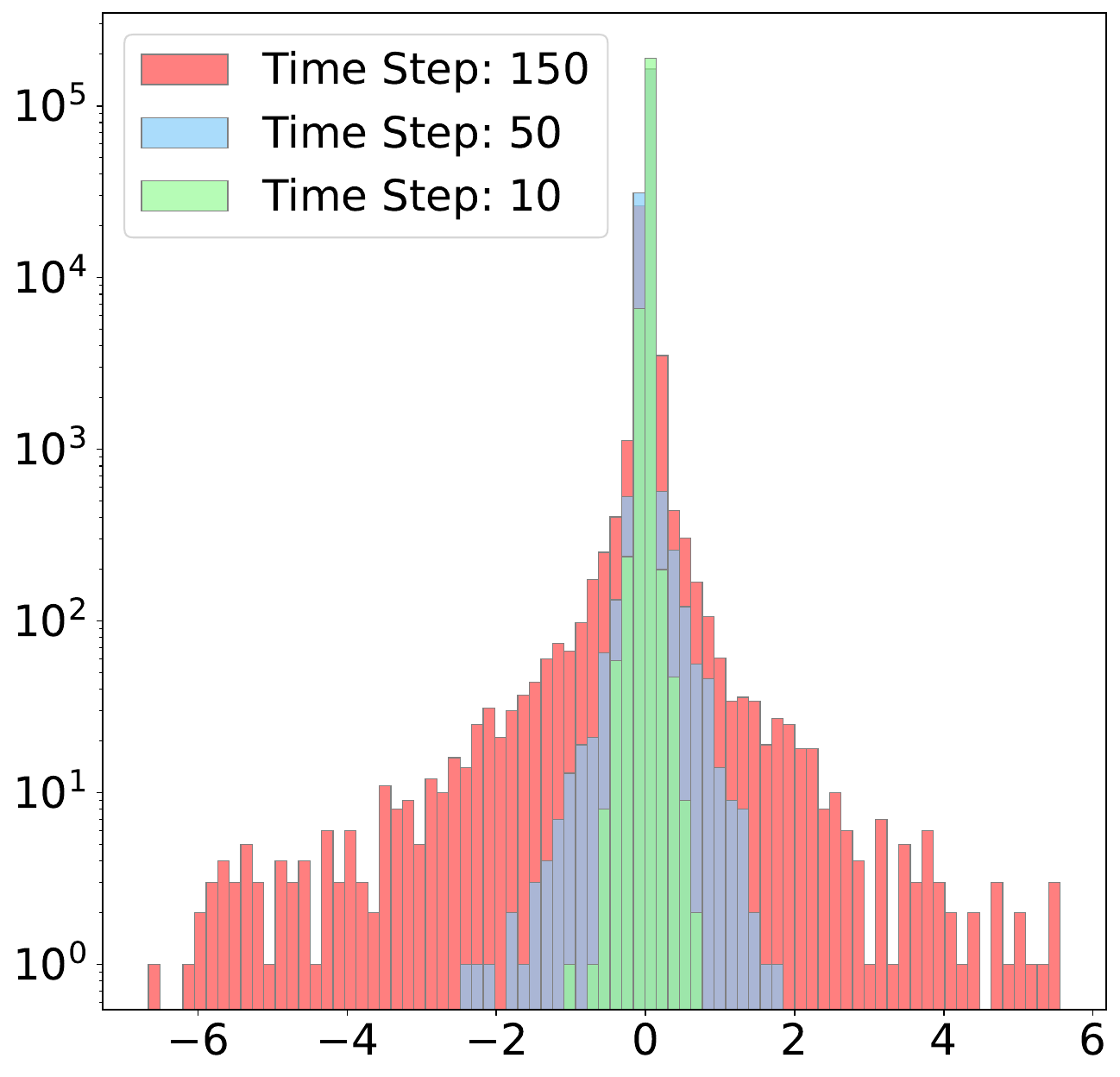}
  }
  \vspace{-3mm}
  \caption{The distributions of discrete parameters $\overline{A}_t$ and hidden states $h_t$, which are a part of the state equation ($h_t=\overline{A}_th_{t-1}+\overline{B}_tx_t$) for the input $x_t$ in the SSM operator. The horizontal axis represents the value range. (a) Long-tailed skewed distribution of discrete parameters $\overline{A}_t$. (b) Highly dynamic variation of hidden states $h_t$. 
 }
  \label{teaser}
\end{figure}

Mamba~\cite{mamba}, a novel and powerful backbone based on state space models (SSMs)~\cite{ssm1,ssm2,ssm3}, has become one of the research hotspots due to its efficient long sequence modeling capability. 
In vision models, SSM-based models~\cite{vim,vmamba,TriPlane,segmamba,mambair} have made impressive progress due to the advanced performance and linear time complexity, which have become a promising alternative to Vision Transformers (ViTs)~\cite{vit,deit}. 
Despite the advantages of SSMs in modeling long sequences, their deployment on various hardware platforms, like edge devices, remains challenging due to the limited memory and power.
Post-Training Quantization (PTQ) is an effective solution for this problem, which can quantize model weights and activations to integers with a limited set of unlabeled calibration datasets, facilitating the deployment of models on resource-limited edge devices with less memory and lower power burden.

However, existing PTQ methods primarily focus on sophisticated optimization strategies~\cite{adaround,brecq, qdrop} or on custom quantizer designs for specific operators (\emph{e.g.}, Softmax in ViTs)~\cite{fqvit,repq,ptq4vit}.  This focus has resulted in a neglect of the quantization analysis for operators within SSMs, creating a void in the availability of quantization methods specially tailored for SSMs.
Since operators of SSMs are different from those of Convolutional Neural Networks (CNNs) and ViTs, including the state equation ($h_t=\overline{A}_th_{t-1}+\overline{B}_tx_t$)~\cite{mamba}, we reveal the quantization sensitivity and outliers in SSM activations by analyzing the activation distribution in SSMs. We notice two distinctive characteristics that pose challenges for the quantization of SSMs: 1) as depicted in Fig.~\ref{teaser:deltaA}, the distribution of the discrete parameters $\overline{A}_t$ within SSMs exhibits long-tailed skewed distributions, with a dense concentration near the maximum value and a sparse distribution at values further from the maximum. This characteristic complicates the process of uniform quantization; 2) as illustrated in Fig.~\ref{teaser:state}, the activation ranges of the hidden states $h_t$ across various time steps in SSMs are highly dynamic. This variability makes it challenging to apply a single quantization parameter across the entire sequence.

In this work, we propose QMamba, one of the first PTQ frameworks tailored for the vision SSMs based on the above observations. First, we propose a Long-tailed Skewness Quantization (LtSQ) to address the long-tailed and skewed distributions of the discrete parameters. LtSQ performs non-uniform quantization for densely distributed activations, ensuring that the multiplication of the quantized discrete parameters and the quantized hidden states can be efficiently implemented as a hardware-friendly bit-shift operation.
Then, we propose a Temporal Group Quantization (TGQ) to handle the dynamic range of hidden state sequences across time steps. 
TGQ groups the hidden state sequences temporally for quantization, enabling fine-grained quantization that adapts to the varying dynamics of the hidden states.


We examine the effectiveness of our QMamba on existing representative SSM-based vision models, \textit{i.e.}, Vim~\cite{vim} and VMamba~\cite{vmamba}, in the image classification task. Through comprehensive experiments across various SSM-based vision models and a range of bit width configurations, we demonstrate that QMamba, optimized for SSM operators, surpasses current PTQ methods in terms of accuracy. Notably, with 4-bit activation values, our method can even outperform existing methods in terms of Top-1 accuracy on the ImageNet classification task with a 21.0\% improvement.

The main contributions of this work are as follows:

1) To the best of our knowledge, QMamba is one of the first PTQ frameworks designed for SSM operators in vision models, filling the gap in quantization methods for SSM operators.

2) We analyze the difficulties of SSM quantization by revealing two distributional characteristics of activation values in SSM based on our observations.

3) We design LtSQ for the long-tailed skewed discrete parameters and TGQ for the highly dynamic hidden states to overcome the challenge of SSM quantization.

4) Extensive experiments demonstrate that our QMamba significantly outperforms existing PTQ methods on representative SSM-based vision models.

\begin{figure*}[t]
  \centering
  \subcaptionbox{Distribution variations of $\overline{A}_t$\label{observation:deltaA_box}}{
    \begin{minipage}{0.235\linewidth}
      \centering
      \includegraphics[width=\linewidth]{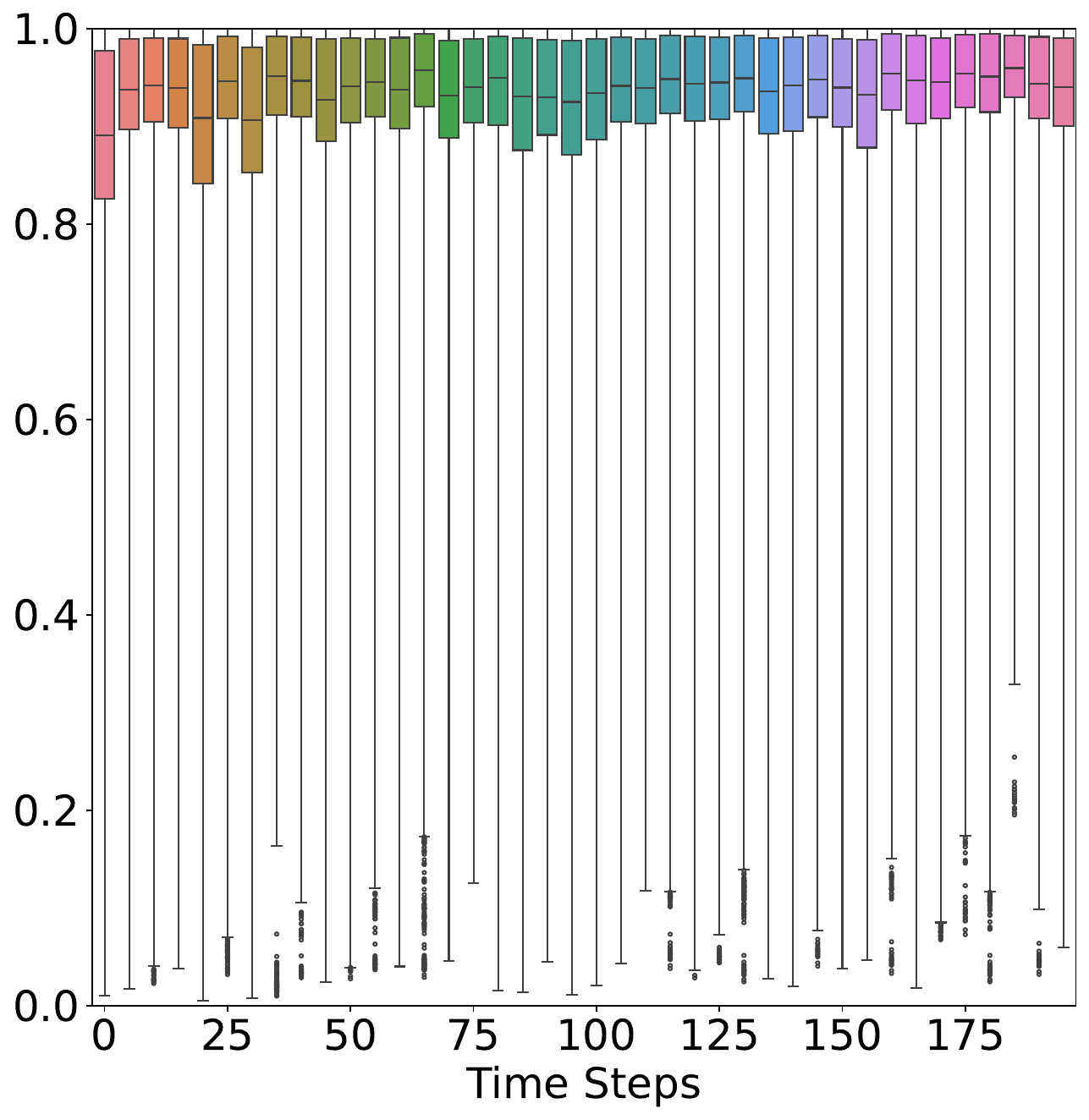}
    \end{minipage}
  }\hfill 
  \subcaptionbox{Distribution variations of $h_t$\label{observation:state_box}}{
    \begin{minipage}{0.25\linewidth}
      \centering
      \includegraphics[width=\linewidth]{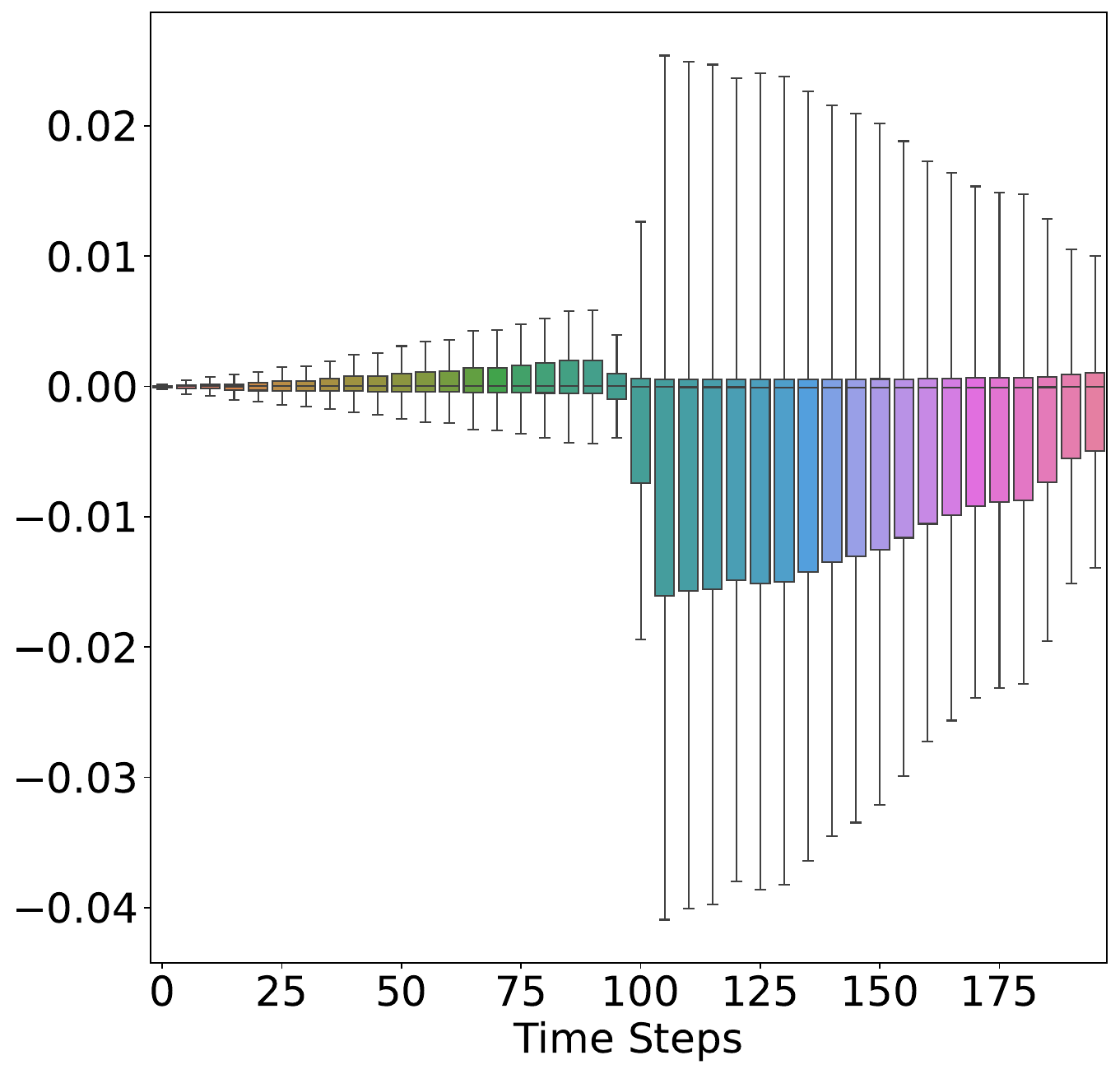}
    \end{minipage}
  }\hfill 
  \subcaptionbox{Uniform quantization on $\overline{A}_t$\label{observation:deltaA_hist}}{
    \begin{minipage}{0.238\linewidth}
      \centering
      \includegraphics[width=\linewidth]{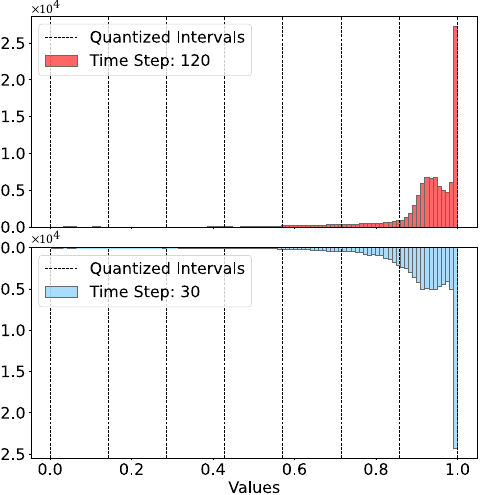}
    \end{minipage}
  }\hfill 
  \subcaptionbox{Uniform quantization on $h_t$\label{observation:state_hist}}{
    \begin{minipage}{0.235\linewidth}
      \centering
      \includegraphics[width=\linewidth]{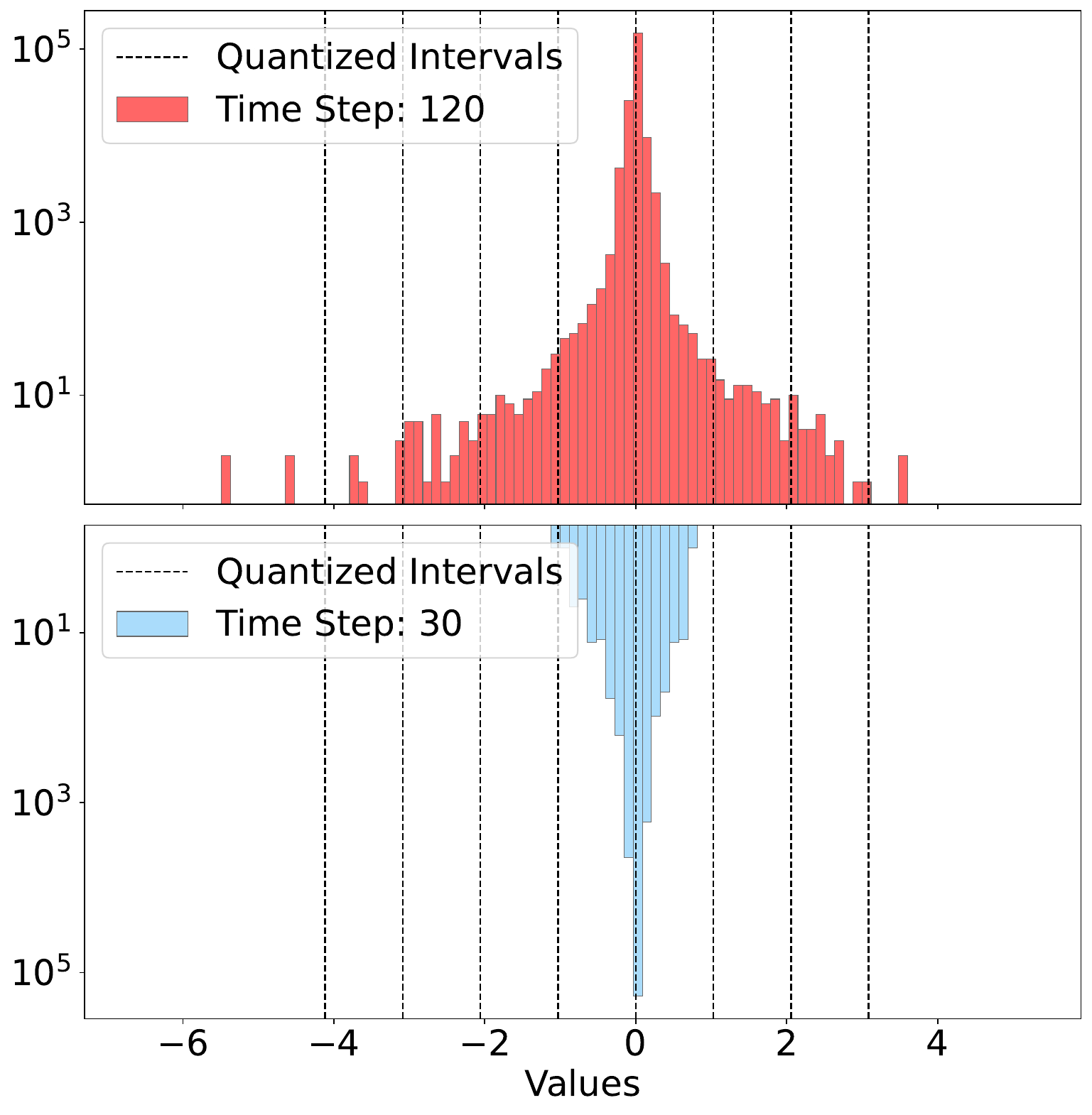}
    \end{minipage}
  }
  \caption{The statistics of discrete parameters $\overline{A}_t$ and hidden states $h_t$ are observed in the SSM of Vim~\cite{vim}. We visualize the distribution range variations of $\overline{A}_t$ and $h_t$ at different time steps in (a) and (b), where the horizontal axis represents the time dimension. For clarity, we visualize $\overline{A}_t$ and $h_t$ at every fifth time step in the sequence, and the outliers in the boxplot of hidden states $h_t$ are omitted for better visualization.
  In (c) and (d), we visualize distribution of $\overline{A}_t$ and $h_t$ at two different time step (\emph{i.e.}, $t=30$ and $t=120$), where the horizontal axis represents the value range. The tensor-wise uniform quantization on $\overline{A}_t$ and $h_t$ at different time steps with a single scaling factor results in uniform and same quantization intervals at different time steps. }
  \label{observation}
\end{figure*}

\section{Related Work}
\label{sec:related}

\subsection{Vision State Space Models}
State Space Models (SSMs) have attracted increasing attention due to their potential for modeling long sequences with linear time complexity. Earlier works based on SSMs~\cite{DBLP:conf/nips/GuJGSDRR21,ssm3,S4D,DSS} have been developed to process sequential data with a focus on capturing long-range dependencies. Building upon these advancements, \citet{mamba} proposes a novel selective SSM, Mamba, which introduces selection mechanisms by incorporating time-varying parameters into the SSM operator, enabling SSMs to selectively propagate or forget information at different time steps of a sequence data. 

For computer vision tasks, recent studies~\cite{vim,vmamba,TriPlane,mambair} have extended SSMs to treat image patches as sequence data to handle spatial dependencies effectively. Vim~\cite{vim} adopts a vision backbone with bidirectional Mamba blocks to model visual representation. VMamba~\cite{vmamba} gathers visual contextual information from multiple perspectives with 2D Selective Scan modules. 
Despite the efficiency of SSMs in long sequence modeling, deploying SSM-based vision models on resource-limited edge devices remains to be explored. Our work aims to provide an effective method for quantization of SSMs on vision tasks, facilitating their deployment on edge devices.


\subsection{Post-Training Quantization}

Quantization is an effective model compression technique that converts weights and activations from floating-point values to low-bit integer values for less memory storage and lower computational consumption. 
Quantization techniques can be broadly categorized into Quantization-Aware Training (QAT)~\cite{pact, lsq, dsq} and Post-Training Quantization (PTQ)~\cite{brecq, qdrop}. QAT jointly optimizes quantization parameters and model weights on labeled datasets, achieving high accuracy but incurring significant training costs. In contrast, PTQ methods focus on optimizing quantization parameters with limited unlabeled calibration datasets, offering a lightweight alternative that avoids the need for retraining. 
OMSE~\cite{OMSE} minimizes the quantization error of weight and activation to achieve low-bit precision inference. Adaround~\cite{adaround} introduces a novel weight-rounding approach that outperforms traditional nearest rounding at low bit widths. BRECQ~\cite{brecq} improves PTQ performance by sequentially reconstructing basic model blocks, achieving comparable performance with QAT methods at 4-bit widths. Qdrop~\cite{qdrop} further enhances PTQ performance by randomly dropping activation quantization in the PTQ process, improving robustness in the quantized model. 
Nevertheless, these advanced methods are not specifically designed for SSMs. The unique operators and dynamics in SSMs present challenges for existing quantization methods. To fill the gap, we propose one of the first PTQ methods for vision SSMs to ensure efficient low-bit quantization by customized design for SSM operators.

\section{Method}
\label{sec:method}

\subsection{Preliminaries}
\textbf{Formulas of Selective SSMs.} 
Selective SSMs introduce a time-varying operator, which maps an input sequence $x_t$ to an output sequence $y_t$ via a sequence of hidden states $h_t$ by the following formulas:
\begin{equation}
\begin{aligned}
  h_t &= \overline{A}_t h_{t-1} + \overline{B}_t x_t, \quad & y_t &= C_t h_t + D x_t, \\
  \overline{A}_t &= \exp{(\Delta_t A)}, \quad & \overline{B}_t &= \Delta_t B_t,
\end{aligned}
\label{equ:selective_ssm}
\end{equation}
where $t$ is the time step (\emph{i.e.}, the $t$-th patch of the images in SSM-based vision models), ($\overline{A}_t$, $\overline{B}_t$) are the discrete parameters, ($A_t$, $B_t$, $C_t$, $D$) are weighting parameters, and $\Delta_t$ is a timescale parameter.
The discrete parameters $\overline{A}_t$, $\overline{B}_t$, and the weighting parameter $C_t$ are dependent on the input $x_t$ as $\Delta_t=Softplus(F_\Delta(x_t))$, $B_t=F_B(x_t)$, and $C_t=F_C(x_t)$. Specifically, $F_B$, $F_C$, and $F_\Delta$ are the linear projection. In this work, we focus on the quantization designed for $\overline{A}_t$ and $h_t$, and we use SSMs as the abbreviation for selective SSMs in the following description.

\noindent\textbf{Formulas of Quantization.} 
Our QMamba is based on the uniform quantization and the log2 quantization. The $b$-bit uniform quantization for a floating-point value $x$ is formulated as follows:
\begin{equation}
\begin{aligned}
  x^q &= clip(\lfloor \frac{x}{s} \rceil + z, 0, 2^b-1), \\
  \hat{x}&=s\cdot(x^q-z)\approx x,
\end{aligned}
\label{equ:uniform_quant}
\end{equation}
where $x^q$ is the value quantized to a $b$-bit integer, and $\hat{x}$ is the de-quantized value approximated to $x$, which can be replaced with the integer $x^q$ in the actual inference~\cite{minmax}. $\lfloor \cdot \rceil$ denotes the round-to-nearest operator, and \emph{clip($\cdot$)} is defined as $clip(x,l,u)=min(max(x, l),u)$. 
The $s$ and $z$ are the scaling factor and the zero point, respectively, both of which are determined by the lower bound $x_{lb}$ and upper bound $x_{ub}$ observed on calibration datasets:
\begin{equation}
\begin{aligned}
  s = \frac{x_{ub}-x_{lb}}{2^b-1}, \quad 
  z = \lfloor -\frac{x_{ub}}{s} \rceil,
\end{aligned}
\label{equ:scale_init}
\end{equation}
For a tensor-wise quantization, both $s$ and $z$ are single scalars used for quantizing an entire tensor of weights or activations.
The log2 quantization is a non-uniform quantization, which is formulated as:
\begin{equation}
\begin{aligned}
  x^q &= clip(\lfloor -\log_2x \rceil, 0, 2^b-1), \\
  \hat{x} &=2^{-x^q}\approx x,
\end{aligned}
\label{equ:log2_quant}
\end{equation}
In this work, we use our LtSQ and TGQ for discrete parameters $\overline{A}_t$ and hidden states $h_t$, respectively, and use tensor-wise uniform quantization for weights and other activations.







\begin{figure}[tp]
  \centering
  \includegraphics[width=0.9\columnwidth]{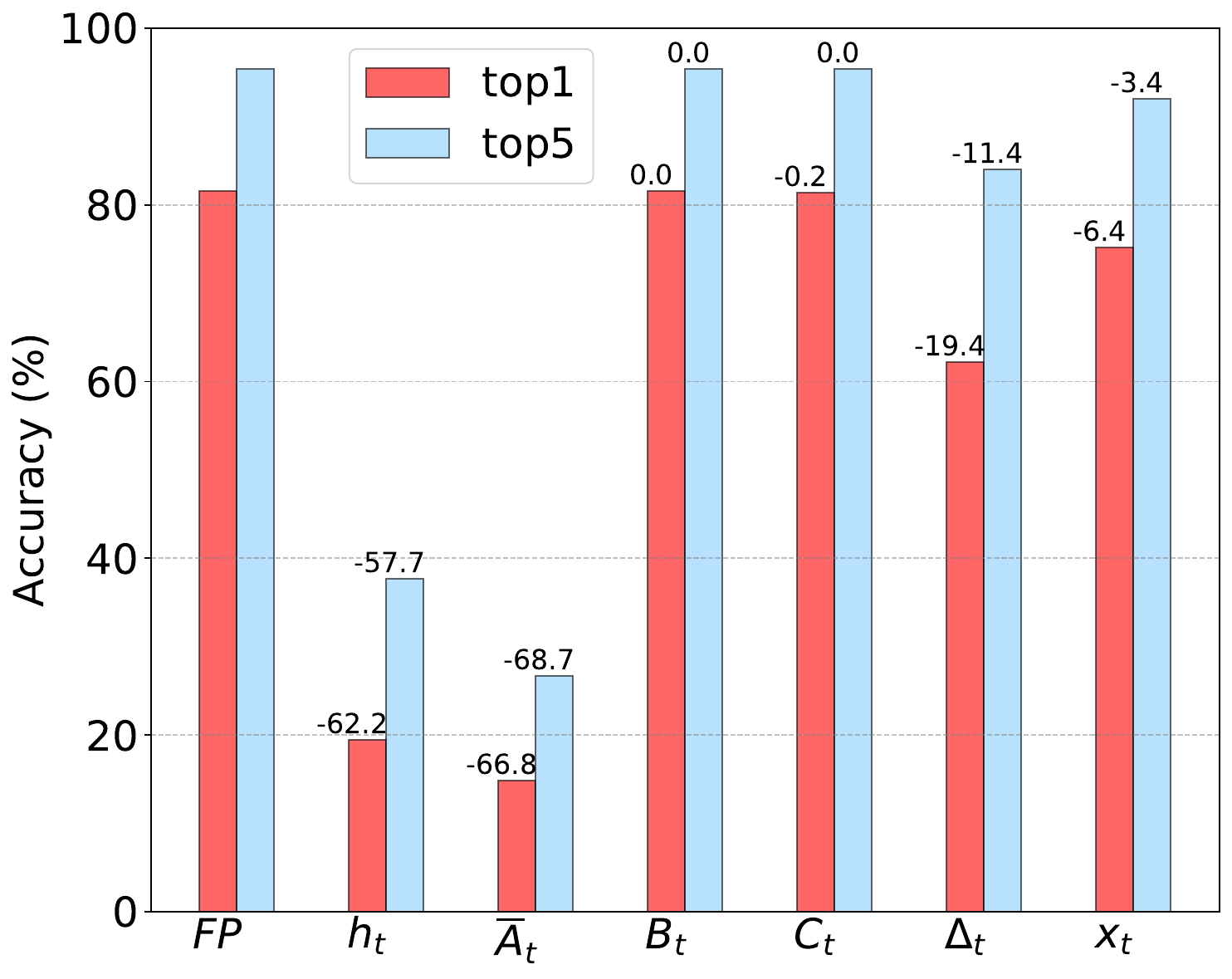}
  \caption{Analysis of quantization sensitivity in different SSM activations. The results report the Top-1 and Top-5 accuracy of Vim-S~\cite{vim} on ImageNet. \textit{FP} denotes the results of the floating-point model. The numbers shown above the bars represent the drop in accuracy compared to \textit{FP}. Each activation is individually quantized to 4 bits.}
  \label{tab:sensi}
\end{figure}

\begin{figure*}[t]
  \centering
  \includegraphics[width=\textwidth]{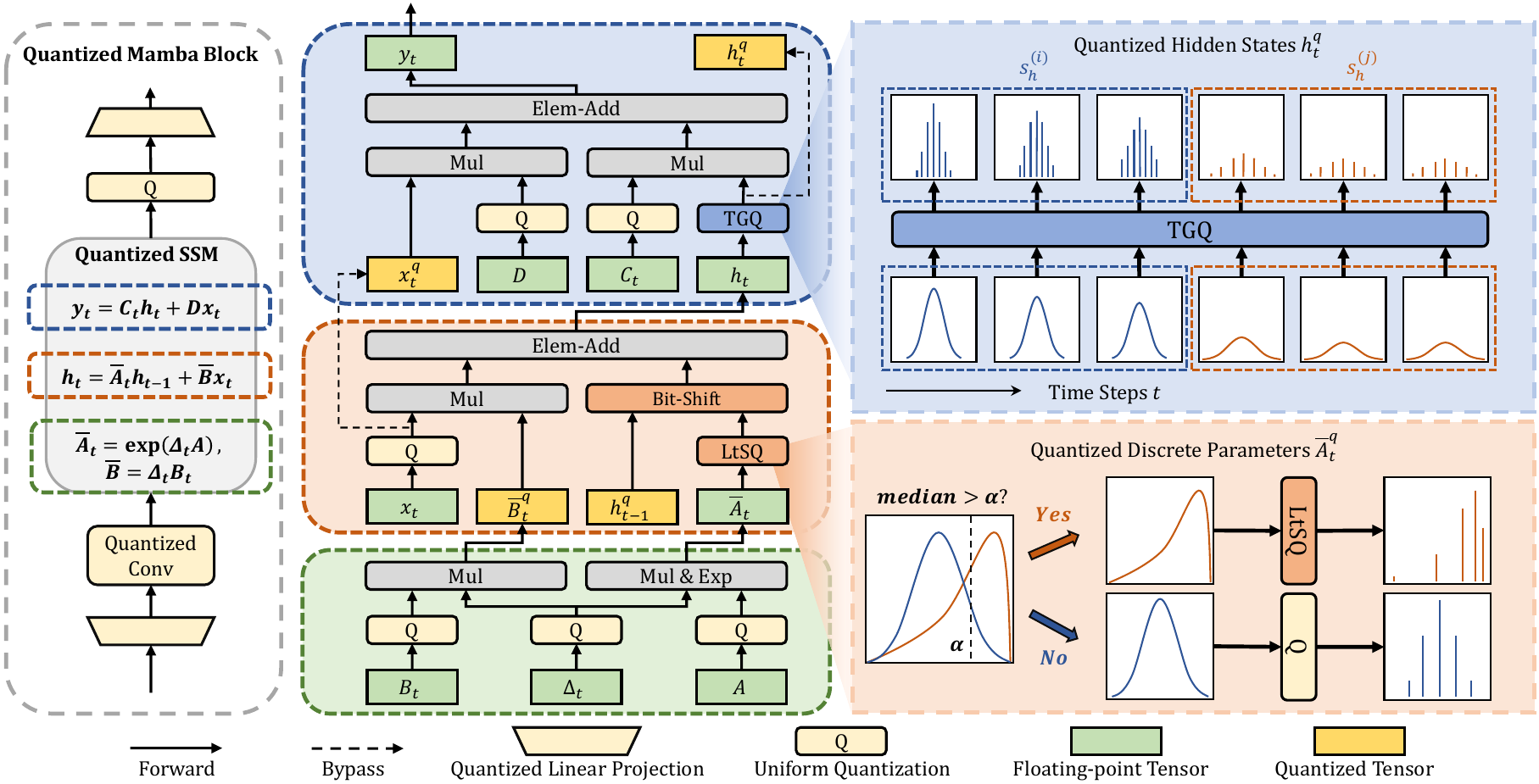}
  \caption{Overview of our QMamba framework. This figure illustrates our quantization framework from the quantized Mamba block to our proposed Long-tailed Skewness Quantization (LtSQ) and Temporal Group Quantization (TGQ) in the quantized SSM. Our LtSQ and TGQ are used for the quantization on discrete parameters $\overline{A}_t$ and hidden states $h_t$, respectively. For a quantized Mamba block, we perform tensor-wise uniform quantization for weights (\emph{i.e.}, $A$ and $D$) and other activations (\emph{i.e.}, $\Delta_t$, $B_t$, $C_t$, and $x_t$) in the SSM operator and for the linear projection and convolution layers in the Mamba block.}
  \label{overview}
\end{figure*}

\subsection{Analysis of Quantization on SSMs}
\label{sec:method:analysis}
In order to explore the quantization sensitivity of SSM activations (\emph{i.e.}, $h_t$, $\overline{A}_t$, $B_t$, $\Delta_t$, $C_t$, and $x_t$), we conduct pre-experiments on each activation of the small version of Vim~\cite{vim} individually to quantize them with the tensor-wise uniform quantization.  
As shown in Fig.~\ref{tab:sensi}, Top-1 accuracy drops 62.2\% and 66.8\% when we quantize hidden states $h_t$ and discrete parameters $\overline{A}_t$ to 4-bit integers individually. When we quantize other activations ($B_t$, $\Delta_t$, $C_t$, and $x_t$) individually, the performance only drops slightly. The pre-experiments reflect the fact that the quantization challenge for SSM results from the activation quantization hidden states $h_t$ and discrete parameters $\overline{A}_t$, inspiring us to design specific quantization methods for SSMs.

As shown in Fig.~\ref{observation}, we further visualize the distribution of discrete parameters $\overline{A}_t$ and hidden states $h_t$ in the SSM operator, and we argue that directly applying tensor-wise uniform quantization to $\overline{A}_t$ and $h_t$ results in a quantization challenge.

\noindent\textbf{Long-tailed skewness of discrete parameters $\overline{A}_t$.} 
As shown in Fig.~\ref{observation:deltaA_box}, the activations of $\overline{A}_t$ at different time steps in the SSM present a dense distribution in a small interval close to 1, with a sparse long tail out of the interval, which we refer to as the long-tail skewness. Some studies have interpreted $\overline{A}_t$ as a forgetting gate in SSMs~\cite{SSM_interpret}, which can decide the decay degree of previous hidden states $h_{t-1}$. Therefore, quantization of dense regions of $\overline{A}_t$ is particularly important, which may affect the capture of long-range dependencies.
As demonstrated in Fig.~\ref{observation:deltaA_hist}, the uniform quantization is suboptimal for discrete parameters $\overline{A}_t$, since quantization intervals are uniform in both sparse and dense regions on the distribution of $\overline{A}_t$, resulting in a coarse quantization on the dense regions of $\overline{A}_t$ with a large quantization error. 



\noindent\textbf{Highly dynamic changes of hidden states $h_t$.}
Different from the distribution of discrete parameters $\overline{A}_t$ shown in Fig.~\ref{observation:deltaA_box}, which has no changes at different time steps, the distribution range of hidden states highly changes at different time steps due to the state equation of the SSM operator.
For example, as shown in Fig.~\ref{observation:state_hist}, the distribution range of the hidden state at the 120-th time step (\emph{i.e.}, $t=120$) is larger than that of the hidden state at the 30-th time step. 
The uniform quantization with a single scaling factor, which determines the interval between quantized values, is not optimal for hidden states $h_t$ with varying distribution ranges. Quantization for hidden states with a large (or small) distribution range may lead to a large quantization error with a small (or large) scaling factor.


Based on the above observations, we propose a PTQ framework, QMamba, to overcome the quantization challenge of vision SSMs. 
As illustrated in Fig.~\ref{overview}, we perform activation quantization for $h_t$, $\overline{A}_t$, $B_t$, $\Delta_t$, $C_t$, and $x_t$, and perform weight quantization for $A$ and $D$ in SSMs, where we use our customized LtSQ and TGQ for discrete parameters $\overline{A}_t$ and hidden states $h_t$, respectively, and use tensor-wise uniform quantization for the other activations and weights.
In the following sections, we will introduce details of our proposed QMamba.

\subsection{Long-tailed Skewness Quantization }
Based on the long-tailed skewed distribution of discrete parameters $\overline{A}_t$, we argue that small quantization intervals are needed in densely distributed regions, while large quantization intervals are needed in sparse long-tailed regions.  
Inspired by the non-uniform quantization for post-softmax activations in ViTs~\cite{ptq4vit, repq}, we propose LtSQ, a tensor-wise non-uniform quantization based on log2 quantization specially designed for discrete parameters $\overline{A}_t$.

As illustrated in Fig.~\ref{overview}, for an accurate activation quantization for discrete parameters $\overline{A}_t$ in a certain SSM, we determine whether the distribution of $\overline{A}_t$ exhibits long-tailed skewness based on a skewness condition:
\begin{equation}
\begin{aligned}
  Median_{1\leq t \leq L}(\overline{A}_t)>\alpha, 
\end{aligned}
\label{equ:ltsq_condition}
\end{equation}
where $\alpha$ is a hyperparameter defined as the skewness boundary, $L$ is the sequence length, and $Median_{1\leq t \leq L}(\cdot)$ represents the median value observed on $\overline{A}_t$ across the entire time steps on the calibration dataset.
If the skewness condition is satisfied, the distribution of $\overline{A}_t$ presents a long-tail skewness, which is not suitable for uniform quantization. In this case, we apply LtSQ quantization; otherwise, we use uniform quantization.
The non-uniform quantization process of LtSQ is designed as:
\begin{equation}
\begin{aligned}
  \overline{A}_t^q &= clip(\lfloor -\log_2(1-\overline{A}_t) \rceil, 0, 2^b-1), \\
  \hat{\overline{A}}_t &=1-2^{-\overline{A}_t^q}\approx \overline{A}_t,
\end{aligned}
\label{equ:LtSQ}
\end{equation}
where $\overline{A}_t^q$ and $\hat{\overline{A}}_t$ are the $b$-bit quantized value and the de-quantized value of $\overline{A}_t$, respectively.
As illustrated in Fig.~\ref{overview}, our LtSQ method applies fine-grained quantization with small intervals for densely distributed regions in the long-tailed skewed distribution, while values in sparse long-tailed regions are quantized with larger intervals.

\begin{table*}[t]
  \caption{Quantitative comparison of different PTQ methods for Vim~\cite{vim} and VMamba~\cite{vmamba} on ImageNet classification. The Top-1 and Top-5 accuracy results of floating-point models are displayed below model names. W8A8, W6A6, and W6A4 represent the bit width of weights and activations, respectively. The \textbf{best results} after PTQ are depicted in bold.}
  \centering
  \label{tab:vim}
  \setlength{\tabcolsep}{1.2mm}{
  \resizebox{\textwidth}{!}{
  \begin{tabular}{c|c|ccccccc}
  \toprule
    \multirow{2}{*}{\makecell{Model \\ Top-1 / Top-5 (\%)}}
    &\multirow{2}{*}{Bit}
        &\multirow{2}{*}{MinMax~\cite{minmax}}
      &\multirow{2}{*}{Percentile~\cite{percentile}} 
      &\multirow{2}{*}{OMSE~\cite{OMSE}} 
      &\multirow{2}{*}{AdaRound~\cite{adaround}}
      &\multirow{2}{*}{BRECQ~\cite{brecq}} 
      &\multirow{2}{*}{QDrop~\cite{qdrop}} 
      &\multirow{2}{*}{\textbf{QMamba~(Ours)}} 
      \\ 
      & & & & & & & & \\ \midrule

\multirow{2}{*}{\makecell{Vim-T \\ 78.3 / 94.2}} 
    &W6A6 &0.1 / 0.7 &18.3 / 36.4 &3.7 / 9.7 &51.2 / 75.4 &51.4 / 75.6 &52.4 / 75.2 &\textbf{57.9 / 82.0}\\
    &W8A8 &1.4 / 3.7 &49.3 / 73.5 &55.5 / 79.5 &57.6 / 81.6 &62.4 / 84.9 &63.9 / 85.6 & \textbf{65.2 / 86.0}\\ \midrule

\multirow{3}{*}{\makecell{Vim-S \\ 81.6 / 95.4}} 
    &W6A4 &0.1 / 0.6 &14.8 / 30.7 &3.5 / 10.8 &21.9 / 34.7 &24.5 / 46.6 &24.9 / 43.3 &\textbf{45.9 / 69.7}\\
    &W6A6 &1.0 / 3.1 &59.5 / 82.4 &36.2 / 61.7 &69.2 / 89.4 &69.7 / 89.8 &70.1 / 89.7 &\textbf{73.1 / 91.0}\\
    &W8A8 &7.5 / 16.6 &64.0 / 86.8 &67.0 / 87.9 &71.3 / 90.9 &71.6 / 90.4 &74.4 / 92.0 &\textbf{77.7 / 93.6}\\ \midrule

\multirow{3}{*}{\makecell{Vim-B \\ 81.9 / 95.8}} 
    &W6A4 &0.1 / 0.6 &47.5 / 70.8 &6.5 / 18.4 &57.8 / 81.5 &60.2 / 83.1 &62.4 / 85.0 &\textbf{65.3 / 86.4}\\
    &W6A6 &0.4 / 1.7 &48.9 / 75.9 &50.9 / 79.1 &66.7 / 86.8 &72.6 / 91.0 &75.2 / 92.7 &\textbf{75.8 / 92.9}\\
    &W8A8 &28.2 / 50.0 &50.3 / 76.6 &56.8 / 86.8 &71.5 / 91.6 &77.2 / 93.4 &78.7 / 94.1 &\textbf{78.9 / 94.3}\\ \midrule  \midrule
    
    \multirow{2}{*}{\makecell{VMamba-T \\ 82.6 / 95.9}} 
    &W6A4 &5.6 / 15.7 &22.3 / 44.5 &21.8 / 43.2 &42.8 / 67.4 &45.6 / 70.6 &51.9 / 77.2 &\textbf{54.9 / 79.2}\\
    &W6A6 &43.3 / 67.3 &63.1 / 83.3 &65.5 / 86.1 &71.0 / 89.4 &77.6 / 93.5 &80.4 / 95.1 &\textbf{80.6 / 95.3}\\ \midrule

\multirow{2}{*}{\makecell{VMamba-S \\ 83.6 / 96.0}} 
    &W6A4 &1.2 / 3.6 &25.2 / 45.1 &10.3 / 21.1 &33.9 / 55.5 &68.2 / 87.5 &69.5 / 88.6 &\textbf{71.5 / 90.2}\\
    &W6A6 &59.7 / 82.6 &72.2 / 90.6 &73.3 / 90.7 &78.2 / 94.1 &80.6 / 95.3 &82.0 / 95.8 &\textbf{82.3 / 96.0}\\ \midrule

\multirow{3}{*}{\makecell{VMamba-B \\ 83.9 / 96.4}} 
    &W6A4 &25.3 / 50.1 &46.5 / 71.1 &46.2 / 71.0 &49.9 / 73.6 &53.7 / 77.3 &56.9 / 78.9 &\textbf{60.0 / 82.1}\\
    &W6A6 &52.7 / 75.6 &74.2 / 90.9 &73.5 / 89.4 &79.0 / 93.4 &81.7 / 95.5 &82.0 / 95.6 &\textbf{82.1 / 95.6}\\
    &W8A8 &77.4 / 93.5 &77.0 / 92.9 &77.3 / 92.5 &81.2 / 95.3 &82.7 / 96.2 &82.9 / 96.2 &\textbf{83.1 / 96.3}\\

\bottomrule
  \end{tabular}
}
}
\end{table*}

\subsection{Temporal Group Quantization}

As mentioned above, since time-varying hidden states $h_t$ are highly dynamic, it is suboptimal to use a tensor-wise uniform quantizer with only a single scaling factor for hidden states at different time steps with a large difference in distribution range.
Here, we propose TGQ to perform a group-wise quantization with different scaling factors for different groups of hidden states $h_t$ in order of time steps. 

As demonstrated in Fig.~\ref{overview}, different from the group-wise quantization in CNNs and ViTs~\cite{group_wise_vit1,group_wise_vit2,group_wise_cnn1}, which divides weights or activations into different groups along the channel dimension and quantizes each group with a separate scaling factor, we group hidden states $h_t$ along the sequence dimension in the order of time steps and apply different scaling factors to each group. 
Suppose the shapes of hidden state sequences $h$ and hidden states $h_t$ at time step $t$ are $(B, L, D, N)$ and $(B, D, N)$, respectively, where $B$ is the batch size, $L$ is the sequence length, $D$ is the expanded state dimension, and $N$ is the SSM dimension~\cite{vim}. 
Specifically, we divide these $L$ hidden states into $\lfloor L/\lambda \rfloor$ groups, where $\lambda$ is a hyperparameter defined as the group length and $\lfloor \cdot \rfloor$ is the floor operator. Then, tensor-wise uniform quantizers with different scaling factors are applied to the corresponding groups of hidden states:

\begin{equation}
\begin{aligned}
  h_t^q &= clip(\lfloor \frac{h_t}{s_h^{(i)}} \rceil + z_h^{(i)}, 0, 2^b - 1), \\
  \hat{h}_t &= s_h^{(i)} \cdot (h_t^q - z_h^{(i)}) \approx h_t, \\
  i &= \min(\lfloor t/\lambda \rfloor, \lfloor L/\lambda \rfloor),
\end{aligned}
\label{equ:TGQ}
\end{equation}
where $i$ is the group index, and $s_h^{(i)}$ and $z_h^{(i)}$ are the scaling factor and zero point applied to the group $i$, respectively. 

\subsection{Quantized Mamba Block}
In order to quantize SSM-based vision models composed of Mamba~\cite{mamba} blocks, we quantize Mamba blocks using our QMamba framework. 
As demonstrated in Fig.~\ref{overview}, after discrete parameters $\overline{A}_t$ and hidden states $h_t$ are quantized using our LtSQ and TGQ, the multiplication in Eq.~\ref{equ:selective_ssm} between $\hat{\overline{A}}_t$ and $\hat{h}_{t-1}$ can be implemented based on a bit shift operator:
\begin{equation}
\hat{\overline{A}}_t\hat{h}_{t-1}=s_h^{(i)}\cdot((h_t^q-z_h^{(i)})-(h_t^q-z_h^{(i)})>>\overline{A}_t^q),
\label{equ:bitshift}
\end{equation}
where $>>$ is the bit shift operator, which makes it a hardware-oriented operation~\cite{fqvit,PTQ4SAM}. 
In addition to the quantized SSM, we also perform quantization on other operators like linear projection and convolution in the Mamba block by weight and activation quantization.

In the PTQ process of our QMamba framework, we first initialize all scaling factors and zero-points of weight and activation quantizers by setting the lower bound $x_{lb}$ and the upper bound $x_{ub}$ in Eq.~\ref{equ:scale_init} to the 1-st and 99-th percentile values observed on calibration datasets, which we will further analyze in detail in Sec.~\ref{sec:scaleinit}. Then, we enable all scaling factors to be learnable and finetune them on the calibration dataset by minimizing the Mean Squared Error (MSE) loss between the output $O_k$ of the $k$-th floating-point Mamba block and the output $\hat{O}_k$ of the quantized Mamba block: 
\begin{equation}
\arg\min_{\mathbf{s}_k} ||O_k-\hat{O}_k||_2,
\label{equ:quantized_mamba}
\end{equation}
where $|| \cdot ||_2$ is the L2 loss function and $\mathbf{s}_k$ represents all scaling factors of the $k$-th Mamba block. The model is fine-tuned block by block, and $\mathbf{s}_k$ are updated through the gradient back-propagation algorithm. More details about the quantized Mamba block and quantized SSM are provided in the supplementary material.

\section{Experiments and Results}
\label{sec:exp}

\subsection{Experimental Setup}

\textbf{Evaluation Settings.} In the evaluation of our QMamba, we conduct experiments on the ImageNet classification task with different bit-width configurations, including W8A8 (8-bit quantization for weights and activations), W6A6 (6-bit quantization), and W6A4 (6-bit quantization for weights and 4-bit quantization for activations). We perform quantization on the representative SSM-based vision models, Vim~\cite{vim} and VMamba~\cite{vmamba}, including their respective tiny, small, and base versions, which we denote as `-T', `-S', and `-B', respectively.
Focusing on activation quantization in SSMs, we set specific bits for some activations of linear projection in Mamba blocks. More details about bit settings can be found in the supplementary material.

\noindent\textbf{Implementation Details.} In the PTQ process, we initialize and optimize scaling factors of weights and activations. Here, we follow the framework of QDrop~\cite{qdrop}. We randomly sample 1024 images from the ImageNet training dataset as the calibration dataset for PTQ and use the ImageNet validation dataset for evaluation. 
We first input the calibration data to the floating-point model to obtain statistics and initialize scaling factors of weights with the observed maximum and minimum values and scaling factors of activations with the observed 1-st and 99-th percentile values. Then, we finetune all initialized scaling factors block by block for 10000 iterations using the Adam optimizer~\cite{adam} with $\beta_1=0.9$ and $\beta_2=0.999$, with a learning rate of $4\times10^{-4}$ and a batch size of 2. The learning rate is scheduled by the CosineAnnealingLR~\cite{CosineAnnealingLR}. We conduct all experiments with PyTorch on Nvidia 3090 GPUs. We set the hyperparameters skewness boundary $\alpha$ of LtSQ and group length $\lambda$ of TGQ to 0.9 and 10 as default when evaluating performance, respectively. The main results are reported as follows.

\noindent\textbf{Baseline Methods.} We select the statistic-based methods (\emph{i.e.}, MinMax~\cite{minmax}, Percentile~\cite{percentile}, and OMSE~\cite{OMSE}) and learning-based methods (\emph{i.e.}, Adaround~\cite{adaround}, BRECQ~\cite{brecq}, and QDrop~\cite{qdrop}) as baseline methods for comparison. For a fair comparison, the learning settings of all learning-based methods are consistent with our QMamba.



\begin{figure*}[t]
  \centering
  \subcaptionbox{LtSQ on $\overline{A}_t$\label{abl:QMamba:ltsq}}{
    \begin{minipage}{0.295\linewidth}
      \centering
      \includegraphics[width=\linewidth]{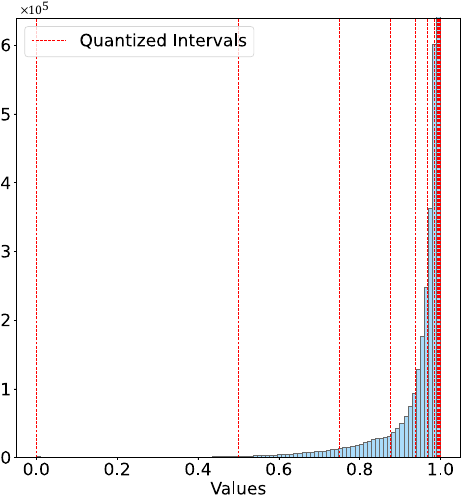}
    \end{minipage}
  }\hfill 
  \subcaptionbox{The variation of $h_t$ and TGQ scaling factors\label{abl:QMamba:tgq1}}{
    \begin{minipage}{0.325\linewidth}
      \centering
      \includegraphics[width=\linewidth]{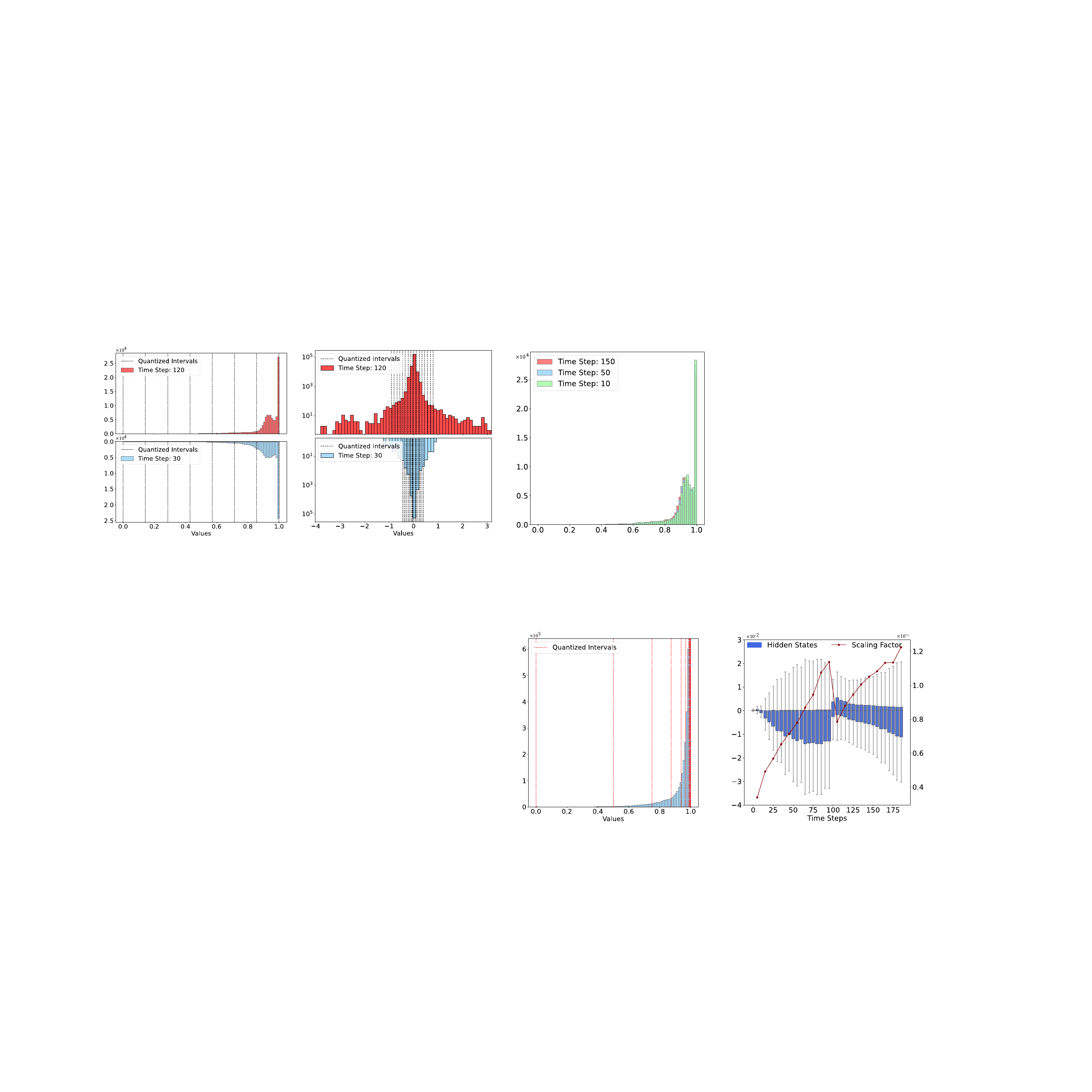}
    \end{minipage}
  }\hfill 
  \subcaptionbox{TGQ on $h_t$\label{abl:QMamba:tgq2}}{
    \begin{minipage}{0.32\linewidth}
      \centering
      \includegraphics[width=\linewidth]{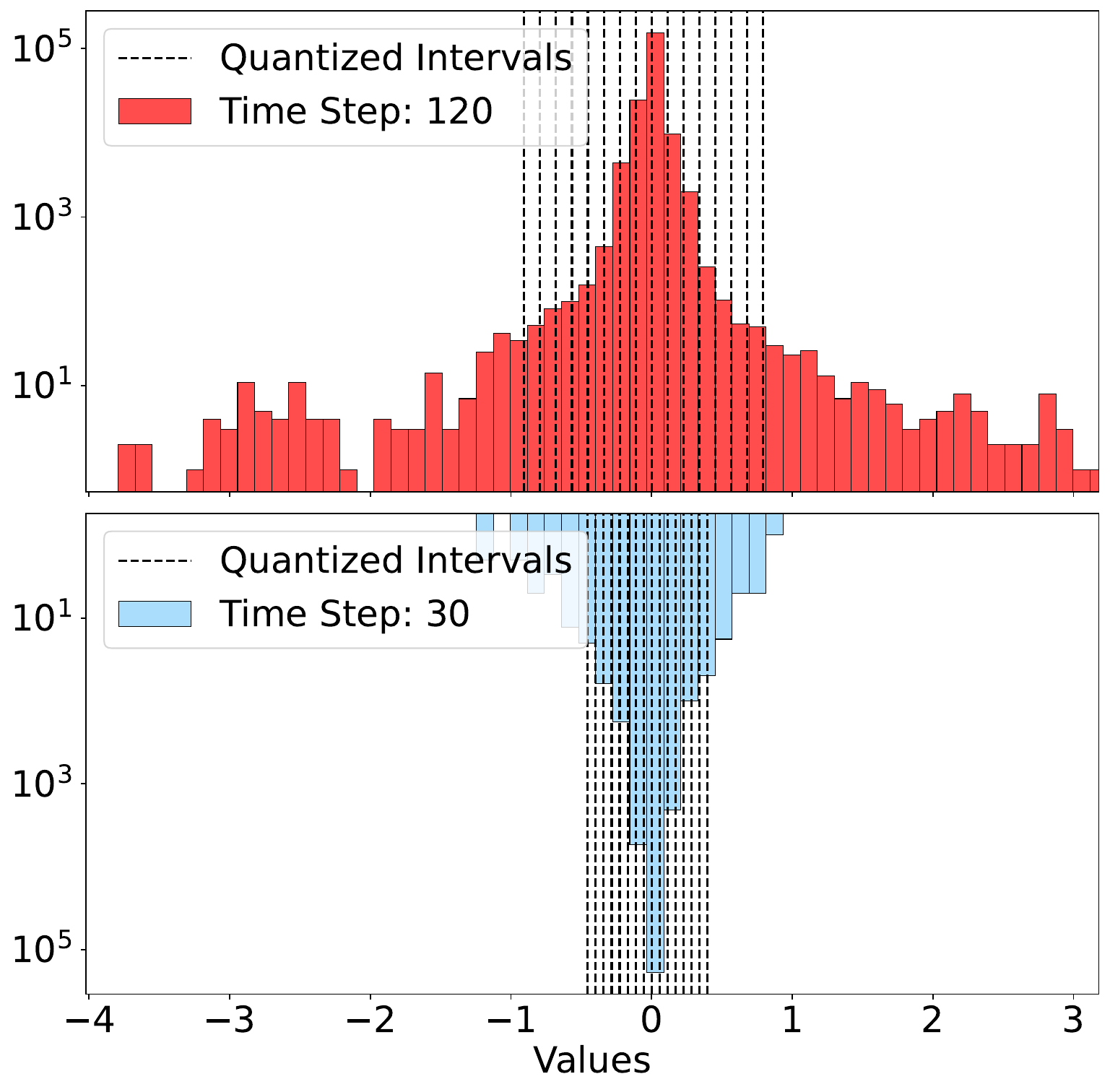}
    \end{minipage}
  }
  \caption{The visualization of our LtSQ and TGQ on discrete parameters $\overline{A}_t$ and hidden states $h_t$ at different time steps in Vim-B (W6A4). (a) The visualization of 4-bit quantization for $\overline{A}_t$ using our LtSQ. (b) The variation of $h_t$ and TGQ scaling factors with time steps. The horizontal axis represents the time step. We show the $h_t$ at every fifth time step and omit outliers in the boxplot for better visualization. (c) The distribution of $h_t$ at different time steps and the corresponding 4-bit quantization intervals using our TGQ. }
  \label{abl:QMamba}
\end{figure*}

\subsection{Comparison Results}
We evaluate our QMamba on multiple quantized versions of Vim and VMamba on the ImageNet classification task. 
As shown in Table~\ref{tab:vim}, our QMamba consistently achieves the highest Top-1 and Top-5 accuracy on Vim and VMamba. We note that the lower the bit-width of the activation values, the more significant advantage our QMamba exhibits over the baseline methods. For example, on Vim-T (W6A6), the Top-1 and Top-5 accuracy of our QMamba is 5.5\%~/~6.8\% higher than that of QDrop (57.9\%~/~82.0\% vs. 52.4\%~/~75.2\%), and on VMamba-T (W6A4), the Top-1 and Top-5 accuracy of our QMamba is 3.0\%~/~2.0\% higher than that of QDrop (54.9\%~/~79.2\% vs. 51.9\%~/~77.2\%). It is worth noting that our QMamba on Vim-S (W6A4) outperforms all baseline methods by a wide margin, \emph{e.g.}, 21.0\%~/~26.4\% higher than QDrop on Top-1 / Top-5 accuracy (45.9\%~/~69.7\% vs. 24.9\%~/~43.3\%). This is because applying tensor-wise uniform quantization to all activations results in a large quantization error. Our proposed LtSQ effectively avoids this issue.

\begin{table}[t]
  \caption{Analysis of different initialization methods for SSM activations of QMamba on ImageNet. All scaling factors are not finetuned. The \textbf{best results} are depicted in bold.}
  \label{tab:init}
  \centering
  \resizebox{\columnwidth}{!}{
  \begin{tabular}{c|ccc|cc}
  \toprule

    \multirow{2}{*}{\textit{Initialization}}
    &\multicolumn{3}{c|}{\textit{Vim-S}}
    &\multicolumn{2}{c}{\textit{VMamba-S}} \\
     &\multicolumn{1}{c}{W8A8} &\multicolumn{1}{c}{W6A6} &\multicolumn{1}{c|}{W6A4} &\multicolumn{1}{c}{W6A6} &\multicolumn{1}{c}{W6A4}\\
      \midrule

MinMax~\cite{minmax} &10.4  &1.2 &0.2 &59.5 &1.2 \\
Percentile~\cite{percentile} &64.0  &\textbf{58.6} &\textbf{29.9} &72.3 &\textbf{29.2} \\
OMSE~\cite{OMSE} &\textbf{68.7}  &35.6 &2.4 &\textbf{73.4} &12.5 \\ 



\bottomrule
  \end{tabular}
  }
\end{table}

\section{Ablation Analysis}
\label{sec:ablation}

\subsection{Initialization of Scaling Factors}
\label{sec:scaleinit}
As discussed in Sec.~\ref{sec:method:analysis}, the activations in SSMs are sensitive to quantization, which suggests that the initialization of scaling factors is important for our QMamba. 
We report the results of initialized models using our QMamba without finetuning any scaling factors in Table~\ref{tab:init}.
We initialize scaling factors with MinMax~\cite{minmax}, Percentile~\cite{percentile}, and OMSE~\cite{OMSE} methods. 
The Percentile initializes scaling factors with the 1-st percentile and 99-th percentile values observed on calibration datasets. Here, we draw three conclusions:
1) MinMax is not a suitable initialization method for activations in SSMs. As listed in Table~\ref{tab:init}, all models that use MinMax to initialize scaling factors perform significantly worse than those initialized using Percentile and OMSE. For example, on Vim-S (W6A6), using MinMax initialization will result in a 57.4\% reduction compared to using Percentile.
2) Percentile is a robust initialization method for activations in SSMs. For example, using Percentil, Vim-S (W6A6) outperforms that using OMSE by 23.0\% and VMamba-S (W6A4) outperforms that using OMSE by 16.7\%.
3) OMSE is not a suitable initialization method for low-bit activation quantization in SSM. When SSM activations are quantized to high bit width, the initialization effect of using OMSE is comparable to that of using Percentile. However, in the case of low bit width, the initialization effect of OMSE is often worse than that of Percentile. The reason is that the initialization process of OMSE is based on minimizing the MSE loss function, which is affected by large-scale outliers in the SSM activations in the low bit width case.



\begin{table}[t]
  \caption{Ablation study on different quantized models with 6-bit weights and 4-bit activations on ImageNet. We report the results of QDrop as the baseline without our customized LtSQ and TGQ. Scaling factors of all models are initialized and finetuned with the same settings. The \textbf{best results} are depicted in bold.}
  \tiny
  \label{tab:ablation}
  \centering
  \resizebox{\columnwidth}{!}{
  \begin{tabular}{cc|ccc}
  \toprule
    \multirow{3}{*}{\textit{LtSQ}}
    &\multirow{3}{*}{\textit{TGQ}}
    &\multicolumn{3}{c}{\textit{Top-1 / Top-5 (\%)}}\\ \cmidrule(lr){3-5}
     &
    &\multicolumn{1}{c}{\textit{Vim-S}} 
    &\multicolumn{1}{c}{\textit{Vim-B}} 
    &\multicolumn{1}{c}{\textit{VMamba-B}} \\
      \midrule

 - &- &24.9 / 43.3  &62.4 / 85.0 &56.9 / 78.9\\
 \ding{51} &- &38.3 / 62.4 &63.9 / 85.5 &58.8 / 81.1\\
    - &\ding{51} &26.3 / 47.7 &64.0 / 85.5 &59.5 / 81.4\\
     \ding{51} &\ding{51} &\textbf{45.9 / 69.7} &\textbf{65.3 / 86.4} &\textbf{60.0 / 82.1}\\ 

\bottomrule
  \end{tabular}
  }
\end{table}

\begin{table}[t]
  \caption{Ablation study on hyperparameters $\alpha$ (skewness boundary) and $\lambda$ (group length). The \textbf{best results} are depicted in bold.}
  \tiny
  \label{tab:ablation_hyper}
  \centering
  \resizebox{\columnwidth}{!}{
  \begin{tabular}{ccccc}
  \toprule
    \multirow{3}{*}{\textit{Row ID}}
    &\multicolumn{2}{c}{\textit{Configuration}}
    &\multirow{1}{*}{\textit{Vim-B}} 
    &\multirow{1}{*}{\textit{VMamba-B}} \\ \cmidrule(lr){2-3}
    
    &\multirow{1}{*}{$\alpha$}
    &\multirow{1}{*}{$\lambda$} &\multicolumn{1}{c}{\textit{W6A4}} &\multicolumn{1}{c}{\textit{W6A4}} \\
      \midrule

1 &0.0 &10 &64.4 / 85.9 &57.0 / 79.8 \\
2 &0.8 &10 &64.9 / 86.0 &59.7 / \textbf{82.4} \\
3 &0.9 &10 &\textbf{65.3 / 86.4} &\textbf{60.0} / 82.1 \\
4 &1.0 &10 &64.0 / 85.5  &59.5 / 81.4 \\ \midrule
5 &0.9 &1 &63.8 / 85.5 &58.1 / 80.7 \\
6 &0.9 &50 &64.5 / 86.1 &57.6 / 80.0 \\

\bottomrule
  \end{tabular}
  }
\end{table}

\subsection{Effectiveness of LtSQ}
The results of Top-1 accuracy listed in Table~\ref{tab:ablation} show the effectiveness of our LtSQ. 
For example, Vim-S, Vim-B, and VMamba-B quantized using our LtSQ alone outperform the corresponding baseline models by 13.4\%, 1.5\%, and 1.9\% on Top-1 accuracy, respectively. When we use LtSQ and our TGQ jointly, the quantized models show higher improvements.
Notably, using our LtSQ and TGQ jointly, Vim-S outperforms the baseline by a significant margin of 21.0\% on Top-1 accuracy (45.9\% vs. 24.9\%). Even when using only our LtSQ, Vim-S achieves a 13.4\% higher accuracy than the baseline (38.3\% vs. 24.9\%). We attribute this to the better initialization performance of the quantized Vim-S when using LtSQ compared to using the uniform quantizer, contributing to the finetuning process of scaling factors.
As shown in Fig.~\ref{abl:QMamba:ltsq}, our LtSQ uses finer-grained quantization intervals for dense areas of long-tailed skewed distributions, which avoids the bad initialization when the discrete parameter $\overline{A}_t$ is quantized to a low bit.
In addition, we conduct experiments with different skewness boundaries $\alpha$ by setting $\lambda=0.9$. As listed in Table~\ref{tab:ablation_hyper}, the best Top-1 results are achieved when $\alpha=0.9$.


\subsection{Effectiveness of TGQ}
We list the results in Table~\ref{tab:ablation} to evaluate the effectiveness of our TGQ. 
For example, the Top-1 accuracy of Vim-B improves by 1.6\% compared to the baseline when using TGQ alone and achieves a higher improvement of 2.9\% when jointly used with LtSQ.
It yields the same conclusion that our TGQ can improve the performance of quantized models, whether used alone or in combination with LtSQ. 

To further analyze the effectiveness of TGQ, we visualize scaling factors of TGQ on Vim in Fig.~\ref{abl:QMamba:tgq1}, which divides hidden states $h_t$ into several groups in time-step order and quantizes them with the corresponding scaling factors, \emph{i.e.}, every red point in Fig.~\ref{abl:QMamba:tgq1} represents a scaling factor of a group. Here, the group length $\lambda$ of TGQ is set to 10. As shown in Fig.~\ref{abl:QMamba:tgq1}, the trends of the scaling factors and the ranges of the hidden state distribution are consistent. 
Furthermore, as shown in Fig.~\ref{abl:QMamba:tgq2}, we visualize quantization intervals of hidden states $h_t$ at two different time steps using our TGQ. The hidden states $h_{120}$ with a large distribution range are quantized with a large interval, while hidden states $h_{30}$ with a small distribution range are quantized with a small interval. Note that the quantization interval is proportional to the corresponding scaling factor, according to Eq.~\ref{equ:scale_init}. It shows that our TGQ can flexibly quantize the dynamically varying hidden states $h_t$ at different time steps.
In addition, we also conduct experiments with different group length $\lambda$ by setting $\alpha=0.9$. As reported in Table~\ref{tab:ablation_hyper} (Row ID: 3, 5, and 6), our TGQ can achieve best results when $\lambda$ is set to 10 for Vim-B and VMamba-B.

\begin{figure}[t]
  \centering
  \includegraphics[width=\columnwidth]{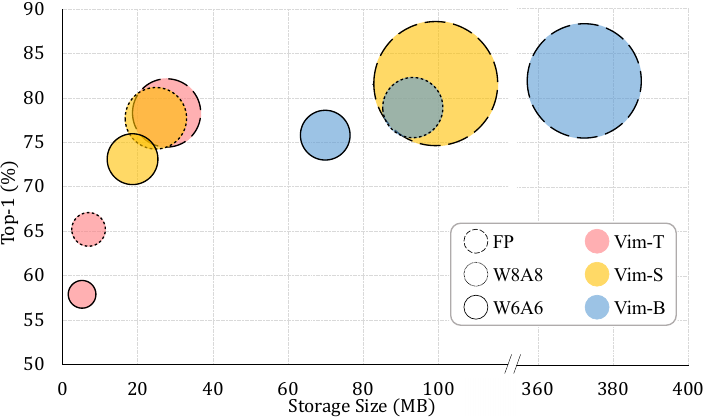}
  \caption{Theoretical efficiency of the quantized Vim. The size of the bubble represents the computational FLOPs.}
  \label{pic:flops}
\end{figure}

\subsection{Theoretical Efficiency Results}
For evaluation on the theoretical efficiency of our QMamba, we follow~\cite{PTQ4SAM,Bi_real_net,vmamba} to calculate the computational FLOPs and storage size of the quantized Vim.
As shown in Fig.~\ref{pic:flops}, quantized models using our QMamba can achieve a better trade-off between performance, storage size, and computational FLOPs compared to floating-point models. For example, QMamba saves up to 80\% of storage size and 75\% of FLOPs when quantizing Vim-B to 6 bits and maintains the high 75.8\% Top-1 accuracy (75.8\% vs. 81.9\%).




\section{Conclusion}
\label{sec:conclusion}

In this work, we first propose QMamba, one of the first PTQ frameworks specifically designed for SSM-based vision models. Our QMamba addresses the quantization challenges posed by the distinctive operator characteristics in SSMs. By analyzing the distributions of discrete parameters and hidden states in SSMs, we identified key challenges in quantizing long-tailed skewed discrete parameters and the highly dynamic hidden states, which we address by introducing LtSQ to handle long-tailed skewed distributions and TGQ to handle the dynamic ranges of hidden states across time steps.
Extensive experiments demonstrate that our QMamba achieves superior results on representative SSM-based vision models, enabling efficient deployment of them on resource-limited edge devices.

\newpage

{
    \small
    \bibliographystyle{ieeenat_fullname}
    \bibliography{main}
}


    














\end{document}